\documentclass[letterpaper, 10 pt, conference]{ieeeconf}  % Comment this line out if you need a4paper
\usepackage[utf8]{inputenc}
\usepackage{graphicx}
\usepackage{amsmath,amssymb,amsfonts}
\usepackage[colorlinks=true, linkcolor=blue, citecolor=blue,anchorcolor=blue, urlcolor=blue]{hyperref}
\usepackage{algorithmic}
\usepackage{cite}
\usepackage{textcomp}
\usepackage{xcolor}
\usepackage{longtable}
\usepackage{mathtools}
\usepackage{multirow} % For multirow cells in tables

\IEEEoverridecommandlockouts                              % This command is only needed if 
 \title{\LARGE \bf
 	Unveiling the Complete Variant of Spherical Robots
 }

 \author{Hassen Nigatu$^{1,2}$, Li Jihao $^{1,2}$, Gaokun Shi$^{1,2}$, Guodong Lu $^{1,3}$, and Huixu Dong$^{*1,2}$% <-this % stops a space
 	\thanks{$^{1}$ Robot Perception and Grasp Lab, School of Mechanical Engineering, Zhejiang University, Hangzhou 310027, China.
 		{\tt\small }}%
 	\thanks{$^{2}$ Robotics Research Center of Yuyao (Robotics Institute of Zhejiang University), Yuyao Technology Innovation Center, No.479 Ye Shan Road , Yuyao, Ningbo Shi, Zhejiang Province, China.
 		{\tt\small }}%
 	\thanks{$^{3}$ The State Key Laboratory of Fluid Power and Mechatronic Systems, School of Mechanical Engineering, and the Engineering Research Center for Design Engineering and Digital Twin of Zhejiang Province, School of Mechanical Engineering, Zhejiang University, Hangzhou 310027, China.
 		{\tt\small huixudong@zju.edu.cn,hassity@gmail.com}}% 
 	\thanks{Correspondence: Huixu Dong, {\tt\small huixudong@zju.edu}.}
 }

\begin{document}

 \newcommand{\join}{\curlyvee}
 \newcommand{\meet}{\curlywedge}

\maketitle

\begin{abstract}
	This study presents a systematic enumeration of spherical ($SO(3)$) type parallel robots' variants using an analytical velocity-level approach. These robots are known for their ability to perform arbitrary rotations around a fixed point, making them suitable for numerous applications. Despite their architectural diversity, existing research has predominantly approached them on a case-by-case basis. This approach hinders the exploration of all possible variants, thereby limiting the benefits derived from architectural diversity. By employing a generalized analytical approach through the reciprocal screw method, we systematically explore all the kinematic conditions for limbs yielding $SO(3)$ motion.Consequently, all 73 possible types of non-redundant limbs suitable for generating the target $SO(3)$ motion are identified. The approach involves performing an in-depth algebraic motion-constraint analysis and identifying common characteristics among different variants. This leads us to systematically explore all 73 symmetric and 5256 asymmetric variants, which in turn become a total of 5329, each potentially having different workspace capability, stiffness performance, and dynamics. Hence, having all these variants can facilitate the innovation of novel spherical robots and help us easily find the best and optimal ones for our specific applications.  
\end{abstract}

%\begin{IEEEkeywords}
%Spherical parallel robots, Screw theory, Complete analysis, Rotation motion generator
%\end{IEEEkeywords}

\section{Introduction}
\begin{figure}[htb!]
	\centering
	\includegraphics[width=\columnwidth]{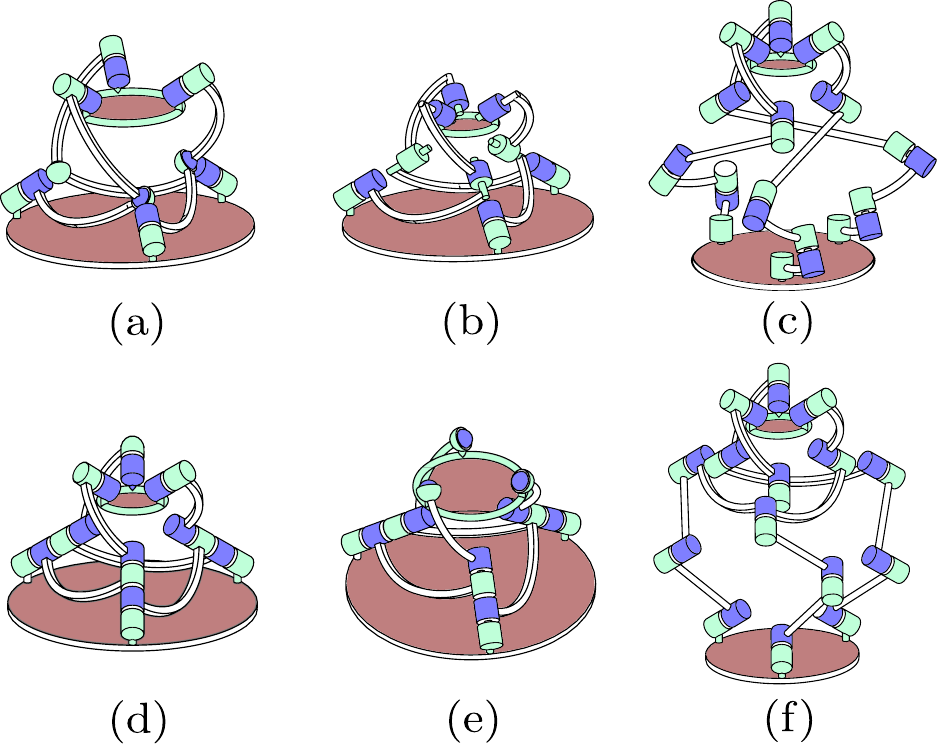} % Adjusts the image to span the column width
	\includegraphics[width=\columnwidth]{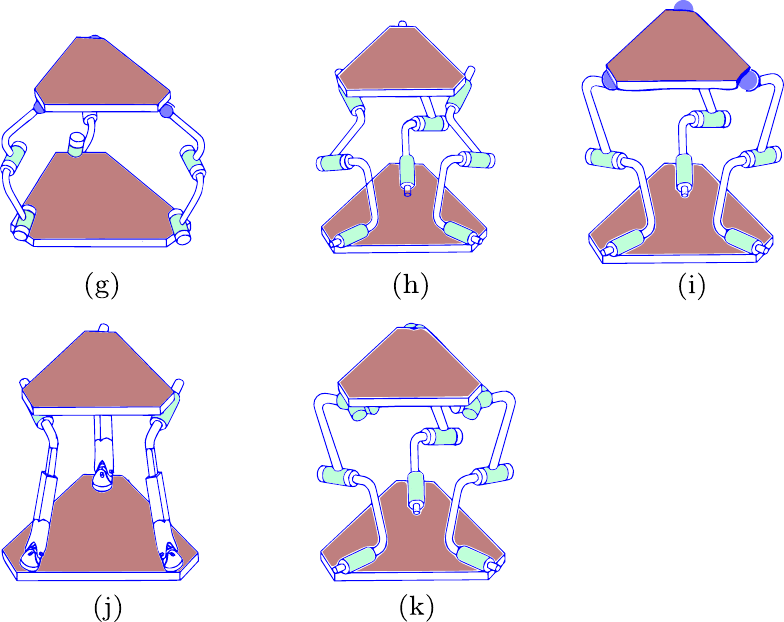}
\caption{Typical examples of spherical parallel robots obtained from Table \ref{tab:limbs_spm}: (a) 3-RSR with all joints' rotation centers meeting at the origin of the moving plate. (b) RCC with all joints' rotation centers meeting at the origin of the moving plate. (c) RRRRR with the last two joints intersecting at the center of the moving plate, while the first three joints intersect at a point other than the center of rotation. (d) 3-RRR intersecting at the center of the moving platform. (e) 3-RRS with all joints' rotation motions passing through the moving platform center. (f) The first three joints are parallel, while the last two joints meet at the center of the moving platform. (g) 3-RRS PM with joints meeting at a point different from the center of the moving platform. (h) 3-CRC with the first and last joints of all limbs meeting at the midpoint of the base plate. (i) 3-CRS with the axes of the first and third joints passing through the center of the base platform. (j) 3-UPC with U and C joints' axes passing through the origin of the base platform. (k) 3-CRU with the first and last joints' axes meeting at the center of the base platform.}
\label{fig:spm}
\end{figure}

Spherical parallel robots, shown in Fig. \ref{fig:spm}, represent a special subset of parallel robots, characterized by their moving platform and links' restriction to rotate about a single point, termed the rotation center \cite{Bai2019}. This distinctive kinematic feature renders them highly advantageous for a broad spectrum of applications. These applications range from the orientation of telescopes, solar panels, and cameras to the realization of wrist, neck, and waist joints for humanoid robots, as well as a multitude of rehabilitation devices, prosthetic robots, and exoskeletons \cite{Bai2019,Chinello2018,Luces2017,Dong2023}. A wealth of research on spherical parallel robotic manipulators has been conducted, encompassing aspects such as dynamics study, workspace modeling, dexterity assessment, design and optimization, singularity analysis, and type synthesis, among other topics \cite{Li2022,Hassani2021,Wei2020,Wu2019,Taunyazov2018,Jin2018,Wu2016,Kong2007,Niyetkaliyev2014,Fang2004,Leguay,Gosselin1989}. The unique capability of these robots to reorient the platform while maintaining its position categorizes them as $SO(3)$ type parallel robots, distinguishing them as an important subset within the realm of robotic manipulators \cite{Meng2007}. Understanding their kinematic, synthesis, and dynamic   

characteristics is fundamental to adapting these robots for specific applications. On the other hand, the $SO(3)$ type parallel robot encompasses a wide variety of limb structures, each offering different workspace ranges, stiffness capabilities, dynamic responses, and other key properties. To select the most suitable robot for a given application, it is crucial to understand all possible varieties and configurations in the most intuitive manner. Thus far, studies on this class of robots have been conducted on a case-by-case basis. However, this method limits our ability to explore the wide range of options based on better performance. 
Typically, these robots generate a 3-DoF spherical motion, denoted as $SO(3)$ within the special Euclidean group $SE(3)$. They can be classified as either serial or parallel types. In the case of parallel robots, which are constructed from a series of limbs, it is essential that each limb incorporates this motion with appropriate constraints to effectively contribute to the overall degrees of freedom (DoF) of the platform once assembled. In this work, we apply an analytical constraint analysis approach to isolated limbs, enabling us to derive a constrained embedded Jacobian. The limb restriction screw (twist of non-freedom) and the limb constraints (wrench of constraint) are elliptic polars \cite{Lipkin1985}. Leveraging this intriguing relationship, we investigate the conditions to identify the relevant direction and moment vectors for the constraint wrenches, which can reveal a variety of limb configurations.       

The remainder of the paper is organized as follows: Section \ref{sec:generic conditions} introduces the very general algebraic conditions required to generate $SO(3)$ type robots. Section \ref{sec:limb_variants} discusses the geometric requirements for $SO(3)$ motion, while Section \ref{sec:limb_variants_2} addresses the specific algebraic conditions that are important for enumerating suitable limbs for this type of robot and provides an enumeration result in Table \ref{tab:limbs_spm}. Finally, the paper concludes in Section \ref{sec:conclusions}.  
%%%%%%%%%%%%%%%%%%%%
\section{Understanding the rate kinematics and relevant conditions} \label{sec:generic conditions}

Generally, the inverse velocity relation for parallel robots can be analytically derived by inverting the extended Jacobian at the limb level, and is as illustrated in (\ref{eq:rik_gen}):

\begin{equation}
	\begin{bmatrix} \dot{\boldsymbol{q}}_a \\ \boldsymbol{0} \end{bmatrix} =  \begin{bmatrix} \boldsymbol{G}_{av}^\top & \boldsymbol{G}_{aw}^\top  \\ \boldsymbol{G}_{cv}^\top & \boldsymbol{G}_{cw}^\top \end{bmatrix} \begin{bmatrix} \boldsymbol{v} \\ \boldsymbol{\omega} \end{bmatrix} = \boldsymbol{G}^\top\boldsymbol{\mathcal{\dot{X}}}\label{eq:rik_gen}
\end{equation} 
where $ \boldsymbol{G} = \begin{bmatrix}  \boldsymbol{G}^\top_{a} \\ \boldsymbol{G}^\top_{c} \end{bmatrix} $

In order to satisfy the structural constraint, $ \boldsymbol{0} = \boldsymbol{G}_{cv}^\top \boldsymbol{v}  +  \boldsymbol{G}_{cw}^\top \boldsymbol{\omega} $ as presented in (\ref{eq:rik_gen}), an arbitrary task velocity, $\boldsymbol{\dot{x}}_a$  can be projected onto the feasible motion space by:

\begin{equation}
	\boldsymbol{\mathcal{\dot{X}}}= \big(\boldsymbol{I} - \boldsymbol{G}_c\boldsymbol{G}_c^\dagger\big)\boldsymbol{\mathcal{\dot{X}}}_a  \label{eq:projection} 
\end{equation}

For spherical manipulators with dependent translational motion, $\boldsymbol{G}_{av}$ has nonzero entries while satisfying the constraint condition, even though it is rank deficient. In this particular case, the matrix $\boldsymbol{G}^\top$ can be inverted as:

\begin{equation}
	\begin{aligned}
		\begin{bmatrix}  \boldsymbol{v} \\ \boldsymbol{\omega} \end{bmatrix} = \begin{bmatrix} \boldsymbol{Q} & \boldsymbol{R} \\ \boldsymbol{S} & \boldsymbol{T} \end{bmatrix}  \begin{bmatrix}  \dot{\boldsymbol{q}}_a \\ \boldsymbol{0} \end{bmatrix} \\ 
	\end{aligned} \label{eq:frk}
\end{equation}
where
\begin{equation*}
	\begin{aligned}
		    \boldsymbol{Q} &=   \boldsymbol{\mathcal{A}}^{-1} + \boldsymbol{\mathcal{A}}^{-1} \boldsymbol{\mathcal{B}}\left(\boldsymbol{\mathcal{D}} - \boldsymbol{\mathcal{C}} \boldsymbol{\mathcal{A}}^{-1} \boldsymbol{\mathcal{B}}\right)^{-1} \boldsymbol{\mathcal{C}} \boldsymbol{\mathcal{A}}^{-1} \\ 
		    \boldsymbol{R} &=  	-\left(\boldsymbol{\mathcal{D}} - \boldsymbol{\mathcal{C}} \boldsymbol{\mathcal{A}}^{-1} \boldsymbol{\mathcal{B}}\right)^{-1} \boldsymbol{\mathcal{C}} \boldsymbol{\mathcal{A}}^{-1} \\ 
		    \boldsymbol{S} &=  -\boldsymbol{\mathcal{A}}^{-1} \boldsymbol{\mathcal{B}}\left(\boldsymbol{\mathcal{D}} - \boldsymbol{\mathcal{C}} \boldsymbol{\mathcal{A}}^{-1} \boldsymbol{\mathcal{B}}\right)^{-1} \\ 
		    \boldsymbol{T} &=  	\left(\boldsymbol{\mathcal{D}} - \boldsymbol{\mathcal{C}} \boldsymbol{\mathcal{A}}^{-1} \boldsymbol{\mathcal{B}}\right)^{-1}
	\end{aligned}
\end{equation*}
and 
\begin{equation*}
	\begin{aligned}
		\boldsymbol{\mathcal{A}} &= \boldsymbol{G}_{av}^\top \\ 
		\boldsymbol{\mathcal{B}} &= \boldsymbol{G}_{aw}^\top \\ 
		\boldsymbol{\mathcal{C}} &= \boldsymbol{G}_{cv}^\top \\ 
		\boldsymbol{\mathcal{D}} &= \boldsymbol{G}_{cw}^\top  
	\end{aligned}
\end{equation*}

Hence, the dependent motion of $\boldsymbol{\mathcal{\dot{X}}}$ becomes $\boldsymbol{v} =  \boldsymbol{Q}\dot{\boldsymbol{q}}_a$ which is different from zero.
If the manipulator has no any dependent translational motion $\boldsymbol{G}_{cv}$ is full rank and $\boldsymbol{G}_{cw} = \boldsymbol{0} $ such that (\ref{eq:rik_gen}) becomes 

\begin{equation}
	\begin{bmatrix} \dot{\boldsymbol{q}}_a \\ \boldsymbol{0} \end{bmatrix} =  \begin{bmatrix} \boldsymbol{G}_{av}^\top & \boldsymbol{G}_{aw}^\top  \\ \boldsymbol{G}_{cv}^\top & \boldsymbol{0}^\top \end{bmatrix} \begin{bmatrix} \boldsymbol{v} \\ \boldsymbol{\omega} \end{bmatrix}   \label{eq:rik_gen_con}
\end{equation}

and the inverse (\ref{eq:rik_gen_con}) can be obtained as:

\begin{equation}
	\begin{bmatrix} \boldsymbol{v} \\ \boldsymbol{\omega}  \end{bmatrix} =  \begin{bmatrix} \boldsymbol{0}   & \boldsymbol{G}_{av}^{-1}   \\ \boldsymbol{G}_{aw}^{-1} & -\boldsymbol{G}_{aw}^{-1}\boldsymbol{G}_{av}\boldsymbol{G}_{aw}^{-1} \end{bmatrix} \begin{bmatrix} \dot{\boldsymbol{q}}_a \\ \boldsymbol{0} \end{bmatrix}   \label{eq:frk_gen_con}
\end{equation}
 
However, from equation  (\ref{eq:frk_gen_con}), it is evident that $\boldsymbol{G}_{av}$ has no effect on relating $\dot{\boldsymbol{q}}$ and $\boldsymbol{\mathcal{\dot{X}}}$. This implies that $\boldsymbol{G}_{av} = \boldsymbol{0}$ is another condition, but not a necessary one, for generating $SO(3)$ type parallel manipulators. Hence, (\ref{eq:rik_gen_con}) and (\ref{eq:frk_gen_con}) are reduced to:

\begin{equation}
	\begin{bmatrix} \dot{\boldsymbol{q}}_a \\ \boldsymbol{0} \end{bmatrix} =  \begin{bmatrix} \boldsymbol{0}^\top & \boldsymbol{G}_{aw}^\top  \\ \boldsymbol{G}_{cv}^\top & \boldsymbol{0}^\top \end{bmatrix} \begin{bmatrix} \boldsymbol{v} \\ \boldsymbol{\omega} \end{bmatrix}   \label{eq:rik_gen_con_final}
\end{equation}

and 
\begin{equation}
	\begin{bmatrix} \boldsymbol{v} \\ \boldsymbol{\omega}  \end{bmatrix} =  \begin{bmatrix} \boldsymbol{0}   & \boldsymbol{G}_{av}^{-1}   \\ \boldsymbol{G}_{aw}^{-1} & -\boldsymbol{0} \end{bmatrix} \begin{bmatrix} \dot{\boldsymbol{q}}_a \\ \boldsymbol{0} \end{bmatrix}   \label{eq:frk_gen_con_final}, \text{respectively.}
\end{equation}    
 
For robots with fewer than 6 degrees of freedom (DoFs), the structural constraint $\boldsymbol{0} = \boldsymbol{G}{cv}^\top \boldsymbol{v} + \boldsymbol{G}{cw}^\top \boldsymbol{\omega}$ is critical and must be satisfied. This condition ensures that the motion complies with the structural constraints.

Now, lets focus on a spherical parallel manipulator that exhibits a pure $SO(3)$ motion, meaning that all translational motions are either entirely eliminated or are retained as dependent motions. From the foundational motion/constraint relation given in (\ref{eq:rik_gen_con}) to (\ref{eq:frk_gen_con_final}), the subsequent \texttt{necessary} and \texttt{sufficient} conditions for the $SO(3)$ manipulator's are derived as follows.

\texttt{Necessary condition}: For the $SO(3)$ manipulator constrained to all linear motions, as indicated by $\boldsymbol{G}_{cw} = \boldsymbol{0}$, the motion of the moving plate must adhere to $\boldsymbol{G}_{cv}^\top \boldsymbol{v} = \boldsymbol{0}$. This constraint results in a filtered motion characterized by $\boldsymbol{v} = (\boldsymbol{I} - \boldsymbol{G}_{vc}\boldsymbol{G}_{cv}^\dagger )\boldsymbol{v}_a$, where $\boldsymbol{v} = \boldsymbol{0}$ and $\boldsymbol{v}_a \neq \boldsymbol{0}$ represent compatible and arbitrary motions, respectively.
 
\texttt{Sufficient condition}: From the forward rate kinematic relations (\ref{eq:frk_gen_con}), it is evident that $\boldsymbol{G}_{av}$ has no influence on $\boldsymbol{v}$, hence $\boldsymbol{G}_{av} = \boldsymbol{0}$. Consequently, this simplifies (\ref{eq:rik_gen_con_final}) to $\dot{\boldsymbol{q}}_a = \boldsymbol{G}_{aw}^\top \boldsymbol{\omega}$.
 
Note that to satisfy the \texttt{Necessary condition}, the limb restriction screw must be $\$_\infty$ screw. If it is a $\$_0$ screw, the direction vector of the wrench becomes zero by the elliptic polarity relation, leading to $\boldsymbol{G}_{cv} = \boldsymbol{0}$ and resulting in a permanent singularity. This can be fundamentally described as follows. 

From the basic screw relation, we have the finite screw representation as:	
	\begin{equation*}
		\$=\left[\begin{array}{c} 		\boldsymbol{s}_\parallel \\ 		\boldsymbol{s}_\parallel \times \boldsymbol{r}  + \lambda \boldsymbol{s} 	\end{array}\right]
	\end{equation*} 
	If the limb restriction twist is $\$_\infty$ pitch, the direction of constraint is along $\boldsymbol{s}_\parallel$, which implies $\boldsymbol{s}_{r\parallel} \neq  \boldsymbol{0} $  while $\boldsymbol{s}_{r\perp} =   \boldsymbol{0} $ to avoid the degenerate case.      

Moreover, the elliptic polarity relationship between a limb restriction screw and the constraint wrench of a limb is stated as follows:   

\textbf{\texttt{Proposition 1:}} From the the polarity of twist and wrench, we know that an $\$_\infty$ limb restriction screw yields a $\$_0$ constraint wrench, i.e.,    

\begin{equation} 
	\begin{aligned}
		\boldsymbol{w}_c &= 
		\begin{bmatrix} 
			\boldsymbol{0} & \boldsymbol{I} \\ \boldsymbol{I} & \boldsymbol{0}	
		\end{bmatrix} \$_r \\ 
	     \begin{bmatrix} 
	     	\boldsymbol{s}_{\parallel} \\\boldsymbol{0}
     	\end{bmatrix} 
     &= \begin{bmatrix} 
	     \boldsymbol{0} & \boldsymbol{I} \\ \boldsymbol{I} & \boldsymbol{0}	
     \end{bmatrix} \begin{bmatrix} \boldsymbol{0} \\ \boldsymbol{s}_{\parallel} \end{bmatrix}
	\end{aligned} 
\label{eq:elliptic_pol} 
\end{equation}  
 
where $\boldsymbol{I} \in \mathbb{R}^{3 \times 3}$ is the identity matrix, $\boldsymbol{w}_c \in \mathbb{R}^6$ represents the constraint wrench, and $\$_r \in \mathbb{R}^6$ denotes the restriction twist screw, respectively. According to the expression in (\ref{eq:elliptic_pol}), $ \boldsymbol{G}_{cv} =  \boldsymbol{0}$ while $ \boldsymbol{G}_{cw} \neq  \boldsymbol{0}$. Which means, for the manipulator to have constrained linear motion, the limb restriction screw must be $\$_\infty$, $\boldsymbol{s}_{r\parallel} \neq \boldsymbol{0}$. On the contrary, if the limb restriction screw $\$_0$, i.e., $\boldsymbol{s}_{r\parallel} = \boldsymbol{0} $ while the moment vector is not. Therefore we can conclude that, $\boldsymbol{s}_{r\parallel} = \boldsymbol{0}$ is the condition for translational manipulators.    

Elliptic polar operator is invariant even after applying the coordinate transformation, i.e.,

\begin{equation}
	\begin{bmatrix} \boldsymbol{w}'_{c\parallel} \\ \boldsymbol{w}'_{c\perp} \end{bmatrix} = \begin{bmatrix}
		\boldsymbol{R} & \boldsymbol{0}  \\  -[\boldsymbol{a}_i]\boldsymbol{R} & \boldsymbol{R}
	\end{bmatrix} \begin{bmatrix} \boldsymbol{0} \\  \boldsymbol{s}_{\parallel}    \end{bmatrix} = \begin{bmatrix} \boldsymbol{0} \\  \boldsymbol{s}'_{\parallel}    \end{bmatrix} \label{eq:trans_wrench}
\end{equation}   

Therefore, the necessary condition for a 3-DoF spherical parallel robot can be more explicitly described as follows: the moment vector of the constraint wrench is zero, which, by the principle of elliptic polarity, implies that the limb's restriction twist is of infinite pitch ($\$_\infty$). Furthermore, it is important to see and consider the nature of the limbs and the overall manipulator from geometric perspective, not only from a pure algebraic standpoint.  

The fundamental requirement for constructing an $SO(3)$  type paralell robot is the presence of at least three legs, each equipped with a single actuator and providing the desired type of three DoFs. Crucially, the combined DoFs from these minimum three legs must be in $SO(3)$ motion space.    

The formal definition of these geometric conditions for the robots under study, which are designed for \(SO(3)\) motion, are as follows:
\[
\left\{
\begin{array}{l}
	l \geq 3,  \text{ is $l$ is the number of legs}, \\
	a_i = 1,  \forall i \in \{1, 2, \ldots, l\}, \text{ $a_i$ is actuators in leg $i$}, \\
	DoF_i >= 3,   \forall i \in \{1, 2, \ldots, L\}, \text{ is the DoF per leg $i$}, \\
	\bigcup_{i=1}^{l} \text{Motion}_i \subseteq SO(3), \text{ is the motion generated by leg $i$}.
\end{array}
\right.
\]

To achieve $SO(3) $ motion in a parallel robot, the motion spaces of all limbs  must intersect, resulting in:

\[
\bigcap_{i=1}^{3} \text{Motion}_i = SO(3)
\]
 Consequently, the limbs that are qualified for generating $SO(3)$ type parallel robots must possess either a 5, 4, or 3 screw system, excluding redundant cases not within the scope of this study. These limbs must adhere to one of the following group based on their screw system:
 
 \begin{itemize}
 	\item For a limb with 5 system, possible classifications include \(5\$0 - 0\$\infty\), \(4\$0 - 1\$\infty\), or \(3\$0 - 2\$\infty\).
 	\item A limb with a 4 system can be categorized as either \(4\$0 - 0\$\infty\) or \(3\$0 - 1\$\infty\).
 	\item A limb with a 3 system is identified as \(3\$0 - 0\$\infty\).
 \end{itemize}
 These categories encapsulate various specific motion generators (limbs) within each, all of which are essential for enabling \(SO(3)\) motion. This necessitates that at least three revolute (R) joints within the manipulator must be independent, facilitating the generation of \(SO(3)\) motion.
 
 In order to fully enumerate these diverse array of limbs belong to this group of manipulator, we implemented the systematic analytic reciprocal screw method \cite{Nigatu2023,Nigatu2021}. The method focuses on analyzing the direction and moment vectors derived of the Jacobian matrix. Subsequent sections will delve into a detailed, element-wise examination of both the motion and constraint screws' direction and moment vectors.
 
\section{Geometric Conditions Needed to be Considered} \label{sec:limb_variants}

Based on geometric analysis  of $SO(3)$ robot design, the following key points are established first for clarity:

\begin{enumerate}
	\item \textbf{Intersection Point Position Invariance:}
	Given two revolute joint axes, \(\boldsymbol{r}_1\) and \(\boldsymbol{r}_2\), on a link intersecting at a point \(\boldsymbol{p}_i\), rotations about these axes by angles \(\theta_1\) and \(\theta_2\), respectively, leave the position of \(\boldsymbol{p}_i\) unchanged. This is mathematically described as:
	\[
	\forall \theta_1, \theta_2, ~ \boldsymbol{R}_{\boldsymbol{r}_1}(\theta_1) \cdot \boldsymbol{p}_i = \boldsymbol{p}_i \quad \text{and} \quad \boldsymbol{R}_{\boldsymbol{r}_2}(\theta_2) \cdot \boldsymbol{p}_i = \boldsymbol{p}_i,
	\]
	where \(\boldsymbol{R}_{\boldsymbol{r}_1}(\theta_1)\) and \(\boldsymbol{R}_{\boldsymbol{r}_2}(\theta_2)\) are the rotation matrices corresponding to rotations about axes \(\boldsymbol{r}_1\) and \(\boldsymbol{r}_2\), respectively. This property signifies the invariant nature of the intersection point under rotations about its defining axes.

	\item \textbf{Parallel Axes Orientation Consistency:}
	This condition ensures that the orientation of parallel joint axes remains consistent during the rotation of a link about any of those axes which can be represented by:
	\[
    \forall \theta_1, \theta_2, \text{ if } \boldsymbol{r}_1 \parallel \boldsymbol{r}_2, \; \text{then}~ \boldsymbol{R}_{\boldsymbol{r}_1}(\theta_1) \parallel \boldsymbol{R}_{\boldsymbol{r}_2}(\theta_2),
    \]	
    \item \textbf{Concurrence Preservation in Intersection:} When two or more consecutive revolute joint axes intersect at a point, the intersection (concurrence) of these axes remains unchanged despite the finite motion of the links.
    \[ 
    \forall \theta_i \text {, the common intersection point of axes } \boldsymbol{r}_i \text { is } \boldsymbol{p}_c \text {, }
    \] where $\boldsymbol{R}_{\boldsymbol{r}_i}(\theta_i)$ represents rotation about axis $\boldsymbol{r}_i$ by angle $\theta_i$, and $\boldsymbol{p}_c$ is the constant point of concurrency for all axes $\boldsymbol{r}_i$.
    \item \textbf{Prismatic Joint Effect:} The inclusion of a prismatic joint among parallel revolute joints does not affect the orientation consistency of the revolute joints.
    \[
    \forall \theta, \text { if } ~ \boldsymbol{r}_1 \| \boldsymbol{r}_2 \text {, then } ~ \boldsymbol{T}_{\text{axis}}(\delta) \cdot \boldsymbol{R}_{\boldsymbol{r}_1}(\theta) \| \boldsymbol{R}_{\boldsymbol{r}_2}(\theta) \text {, }
    \] where $\boldsymbol{T}_{\text{axis}}(\delta)$  represents the translation along the prismatic joint and is the $\delta$ displacement.
\end{enumerate}

Furthermore, $\$_0$ joints are categorized based on their spatial relationship to the moving platform's center of rotation as follows:

\begin{enumerate}
	\item \textbf{Category A (Intersecting at the Center):} Joints for which the vector $\boldsymbol{r}_i$ defining the axis of rotation intersects the center of rotation $\boldsymbol{c}$, i.e., $\boldsymbol{r}_i \curlyvee \boldsymbol{c} \neq \varnothing$, where $\curlyvee$ depicts intersection.
	
	\item \textbf{Category B (Not Intersecting at the Center):} Joints where the rotation axis vector $\boldsymbol{r}_i$ does not intersect the center of rotation $\boldsymbol{c}$, i.e., $\boldsymbol{r}_i \curlyvee \boldsymbol{c} = \varnothing$. A prismatic joint, defined by translation vector $\mathbf{t}$, may be integrated with these joints, not affecting their classification.
\end{enumerate}

For a PM, Category A $\$_0$ joints are defined by the following constraints:

\begin{enumerate}
	\item To uniquely determine a center of rotation $\boldsymbol{c}$ for the moving platform, each limb must include at least two Category A $\$_0$ joints, represented by axes vectors $\boldsymbol{r}_1$ and $\boldsymbol{r}_2$, which intersect at point $\boldsymbol{c}$.
	
	\item The maximal number of Category A $\$_0$ joints per limb is three, dictated by the requirement for ${\boldsymbol{r}_1, \boldsymbol{r}_2, \boldsymbol{r}_3}$ to intersect at a singular point $\boldsymbol{c}$, thereby defining a spherical motion space.
	
	\item If a limb contains more than three Category A $\$_0$ joints, indicated by $\{\boldsymbol{r}_1, \boldsymbol{r}_2, \boldsymbol{r}_3, \ldots\}$, it implies redundancy among the joints, as their axes will also intersect at the same point $\boldsymbol{c}$. Such cases are not in the scope of this paper due to the redundancy.
	
	\item Except for a specific limb configuration referred to as ideal $SO(3)$ parallel robot, which comprises exactly three Category A $\$_0$ joints intersecting at $\boldsymbol{c}$, a limb should ideally have exactly two Category A $\$_0$ joints. These joints must either be adjacent or located at the two extremities of the limb to effectively define the spherical motion space.
\end{enumerate}

Furthermore, the Category-B $\$_0$  and $\$_\infty$  joints can be defined as:

Let the center of rotation be denoted as $\boldsymbol{c}$.
Category-A $\$_0$ joints' contribution to linear velocity at $\boldsymbol{c}$ is zero, represented as $\boldsymbol{v}_c = \boldsymbol{0}$.
With Category-A joints fixed, Category-B $\$_0$ and $\$_\infty$ joints induce rotation about $\boldsymbol{c}$.
The total number of Category B $\$_0$ and $\$_\infty$ joints required to form this linkage is three.
For all-$\$_0$ configurations, the axes conditions are $\boldsymbol{r}_i \parallel \boldsymbol{r}_j$ or $\boldsymbol{r}_i \curlywedge \boldsymbol{r}_j \neq \boldsymbol{c}$, $i \neq j$, where $\curlywedge $ depicts intersection.
A prismatic ($\$_\infty$ ) joint, if present, does not alter these conditions and can be placed amongst parallel $\$_0$ joints: $\$_\infty \in {\boldsymbol{r}_i}$.

Leveraging the above detailed geometric insights into the $SO(3)$ type parallel robots, subsequent sections will further elucidate and establish the conditions through a purely algebraic approach.

%%%%%%%%%%%%%%%%%
\section{Specific Conditions for All Limb Variants} \label{sec:limb_variants_2}

\subsection{The $5\$_0$ system limbs} \label{subsec:five system limbs}
The general constraint equation for the $5\$_0$ limb system is 

\begin{equation}
	\begin{bmatrix}
		\boldsymbol{s}_{5\parallel} &  (\boldsymbol{s}_{5\parallel} \times \boldsymbol{l}_5)^\top \\ 
		\boldsymbol{s}_{4\parallel} &  (\boldsymbol{s}_{4\parallel} \times \boldsymbol{l}_4)^\top \\
		\boldsymbol{s}_{3\parallel} &  (\boldsymbol{s}_{3\parallel} \times \boldsymbol{l}_3)^\top \\
		\boldsymbol{s}_{2\parallel} &  (\boldsymbol{s}_{2\parallel} \times \boldsymbol{l}_2)^\top \\
		\boldsymbol{s}_{1\parallel} &  (\boldsymbol{s}_{1\parallel} \times \boldsymbol{l}_1)^\top \\
	\end{bmatrix}   \begin{bmatrix} \boldsymbol{s}_{r\perp} \\ \boldsymbol{s}_{r\parallel} \end{bmatrix} = \begin{bmatrix} \boldsymbol{C}_1 & \boldsymbol{C}_2 \end{bmatrix} \begin{bmatrix} \boldsymbol{s}_{r\perp} \\ \boldsymbol{s}_{r\parallel} \end{bmatrix} =\boldsymbol{0}  \label{eq:five_zero_sys}
\end{equation}
Here, from $\boldsymbol{C}_1\boldsymbol{s}_{r\perp} +  \boldsymbol{C}_2\boldsymbol{s}_{r\parallel} = \boldsymbol{0}$ we can understand that $\boldsymbol{C}_1$ should be full rank to have $\boldsymbol{s}_{r\perp} = \boldsymbol{0}$. Furthermore, $\boldsymbol{C}_2$ should be rank deficient to have $\boldsymbol{s}_{r\parallel} \neq \boldsymbol{0}$. 
 
Solving (\ref{eq:five_zero_sys}) for $\boldsymbol{s}_{r\parallel}$ yields: 
\begin{equation}
	\begin{aligned}
		\boldsymbol{s}_{r \|}= 
		& -\left[\boldsymbol{s}_{5  \|}^\top\left(\boldsymbol{s}_{4  \|} \times \boldsymbol{s}_{3  \|}\right)\right]\left(\left(\boldsymbol{l}_2 \times \boldsymbol{s}_{2  \|}\right) \times\left(\boldsymbol{l}_{1} \times \boldsymbol{s}_{1  \|}\right)\right) \\
		& +\left[\boldsymbol{s}_{5 \|} ^\top \left(\boldsymbol{s}_{3 \|} \times \boldsymbol{s}_{1  \|}\right)\right]\left(\left(\boldsymbol{l}_4 \times \boldsymbol{s}_{4  \|}\right) \times\left(\boldsymbol{l}_{2} \times \boldsymbol{s}_{2  \|}\right)\right) \\
		& -\left[\boldsymbol{s}_{5  \|} ^\top\left(\boldsymbol{s}_{4 \|} \times \boldsymbol{s}_{1 \|}\right)\right]\left(\left(\boldsymbol{l}_3 \times \boldsymbol{s}_{3 \|}\right) \times\left(\boldsymbol{l}_{2} \times \boldsymbol{s}_{2 \|}\right)\right) \\
		& +\left[\boldsymbol{s}_{5 \|} ^\top\left(\boldsymbol{s}_{3  \|} \times \boldsymbol{s}_{2  \|}\right)\right]\left(\left(\boldsymbol{l}_4 \times \boldsymbol{s}_{4  \|}\right) \times\left(\boldsymbol{l}_{1} \times \boldsymbol{s}_{1  \|}\right)\right) \\
		& -\left[\boldsymbol{s}_{5 \|} ^\top\left(\boldsymbol{s}_{4  \|} \times \boldsymbol{s}_{1  \|}\right)\right]\left(\left(\boldsymbol{l}_4 \times \boldsymbol{s}_{4  \|}\right) \times\left(\boldsymbol{l}_{3} \times \boldsymbol{s}_{3  \|}\right)\right) \\
		& +\left[\boldsymbol{s}_{5  \|} ^\top\left(\boldsymbol{s}_{4  \|} \times \boldsymbol{s}_{2  \|}\right)\right]\left(\left(\boldsymbol{l}_3 \times \boldsymbol{s}_{3  \|}\right) \times\left(\boldsymbol{l}_{1} \times \boldsymbol{s}_{1  \|}\right)\right)
	\end{aligned} \label{eq:five_zero_direc}
\end{equation}
	
Based on the geometric insights into $SO(3)$ type parallel robots, we cannot directly analyze the conditions leading to $\boldsymbol{s}_{r\perp} = \boldsymbol{0}$ from (\ref{eq:five_zero_direc}). However, it is straightforward to obtain $\boldsymbol{s}_{r\parallel} = \boldsymbol{0}$, which is a necessary condition for the translational manipulator, i.e.
	
	\begin{itemize}
		\item Four direction vectors are parallel: For instances; if $\boldsymbol{s}_{1 \|}$, $\boldsymbol{s}_{2 \|}$, $\boldsymbol{s}_{3 \|}$, and $\boldsymbol{s}_{4 \|}$ (\texttt{not applicable}). 
		
		\item Three parallel direction vectors combined with parallelism between a position vector and a direction vector.
		
		\item Two parallel direction vectors combined with parallelism between position vectors and direction vectors.
	\end{itemize}

Then, by duality, the conditions for $\boldsymbol{s}_{r\perp} = \boldsymbol{0}$ can be easily obtained. Here, duality refers to the notion that a rotational manipulator serves as the dual counterpart of a translational manipulator within the $SE(3)$ structure.

Consequently, (\ref{eq:five_zero_direc}) provides the following detailed conditions for nullifying the limb restriction twist's moment vector, ensuring $\boldsymbol{G}_{cw} = \boldsymbol{0}$.
In line with earlier discussions, a 3-DoF spherical parallel robot necessitates the presence of three independent rotational axes for each limb to ensure full spherical motion capability. These axes are denoted by $\boldsymbol{s}_{i\parallel}$, $(\boldsymbol{s}_{j\parallel} \neq \boldsymbol{s}_{i\parallel})$, and $\boldsymbol{s}_{i\parallel} \times \boldsymbol{s}_{j\parallel}$, guaranteeing their independence. Based on this fact and above generic conditions, the following two valid cases are obtained for $5\$_0$ limb 
 
 \begin{itemize}
 	\item  \texttt{Case 1:} 2R joints axes on $\parallel$ planes.
 	\item  \texttt{Case 2:} 3R joints axes on $\parallel$ planes .
 \end{itemize}
 
 To systematically enumerate limbs falling under \texttt{Case 1}, the \texttt{Necessary} and \texttt{Sufficient conditions} are outlined as follows. 
 
\subsubsection{The $5\$_0$ system limbs \texttt{Case 1}} 

To obtain limb variants of the $5\$_0$ screw system where two revolute joints are parallel, the following general equation is established: 

\begin{equation}
	\begin{bmatrix}
		\boldsymbol{s}_{1\parallel} &  (\boldsymbol{s}_{1\parallel} \times \boldsymbol{l}_1)^\top \\ 
		\boldsymbol{s}_{1\parallel} &  (\boldsymbol{s}_{1\parallel} \times \boldsymbol{l}_2)^\top \\
		\boldsymbol{s}_{3\parallel} &  (\boldsymbol{s}_{3\parallel} \times \boldsymbol{l}_3)^\top \\
		\boldsymbol{s}_{4\parallel} &  (\boldsymbol{s}_{4\parallel} \times \boldsymbol{l}_4)^\top \\
		\boldsymbol{s}_{5\parallel} &  (\boldsymbol{s}_{5\parallel} \times \boldsymbol{l}_5)^\top \\
	\end{bmatrix}   \begin{bmatrix} \boldsymbol{s}_{r\perp} \\ \boldsymbol{s}_{r\parallel} \end{bmatrix} = \begin{bmatrix}  \boldsymbol{C}_1 & \boldsymbol{C}_2  \end{bmatrix}  \begin{bmatrix} \boldsymbol{s}_{r\perp} \\ \boldsymbol{s}_{r\parallel}  \end{bmatrix} =  \boldsymbol{0}  \label{eq:case II_con}
\end{equation} where $\boldsymbol{s}_{5\parallel} = \boldsymbol{s}_{1\parallel} \times \boldsymbol{s}_{3\parallel}$.

Solving the above constraint equation gives the direction vector with the following two sub-cases:  

I) The direction vector of the restriction twist $\boldsymbol{s}_{r\parallel} = \boldsymbol{s}_{1\parallel}$, under the conditions $\boldsymbol{s}_{1\parallel} \times \boldsymbol{l}_{12} = \boldsymbol{0}$ or $\boldsymbol{s}_{3\parallel} \times \boldsymbol{l}_{12} = \boldsymbol{0}$, leads to a null vector. Consequently, $\boldsymbol{C}_2\boldsymbol{s}_{r\parallel} = \boldsymbol{0}$, yielding the following conditions. In this context, $\boldsymbol{l}_{ij}$ represents the position vector between joints $i$ and $j$.

\begin{enumerate}
	\item $ \boldsymbol{s}_{1\parallel} \times  \boldsymbol{l}_{3} = \boldsymbol{0} \vee \boldsymbol{s}_{3\parallel} \times   \boldsymbol{l}_{3} = \boldsymbol{0} $
	\item $ \boldsymbol{s}_{3\parallel} \times \boldsymbol{l}_{4} = \boldsymbol{0} \vee \boldsymbol{s}_{1\parallel} ) \times \boldsymbol{l}_{4} = \boldsymbol{0} $
	\item $ \boldsymbol{s}_{3\parallel}^\top\boldsymbol{l}_{5} \wedge  (\boldsymbol{s}_{1\parallel} ^\top \boldsymbol{l}_5 \vee \boldsymbol{s}_{3\parallel}^\top \boldsymbol{s}_{1\parallel} = \boldsymbol{0})  $
\end{enumerate}

II) $ \boldsymbol{s}_{r\parallel} = \boldsymbol{l}_{12}$ with $ \boldsymbol{s}_{3\parallel} \times  \boldsymbol{l}_{12} = \boldsymbol{0}  \vee ( \boldsymbol{s}_{3\parallel} \times \boldsymbol{l}_{34} = \boldsymbol{0} \vee \boldsymbol{l}_{12} \times  \boldsymbol{l}_{34} = \boldsymbol{0})$ 
where $\wedge$ and $\vee$ represent ``and" and ``or", respectively.

The conditions to have zero moment vector ($\boldsymbol{s}_{r\perp} = \boldsymbol{0} $) which should be simultaneously satisfied are: 

\begin{enumerate}
	\item $  \boldsymbol{l}_1 \times  \boldsymbol{s}_{1\parallel} = \boldsymbol{0} \vee \boldsymbol{l}_1 \times \boldsymbol{l}_2 = \boldsymbol{0} $
	\item $  \boldsymbol{s}_{3\parallel} \times \boldsymbol{l}_3 = \boldsymbol{0} \vee  \boldsymbol{s}_{3\parallel} \times \boldsymbol{l}_{12} = \boldsymbol{0}  \vee  \boldsymbol{l}_3  \times  \boldsymbol{l}_{12} = \boldsymbol{0}$
	\item $  \boldsymbol{s}_{3\parallel} \times \boldsymbol{l}_4 = \boldsymbol{0}  \vee \boldsymbol{s}_{3\parallel} \times  \boldsymbol{l}_{12} = \boldsymbol{0}  \vee  \boldsymbol{l}_{4}  \times \boldsymbol{l}_{12} = \boldsymbol{0} $
	\item $  \boldsymbol{l}_5 ^\top  \boldsymbol{s}_{3\parallel} = \boldsymbol{0}  \wedge ( \boldsymbol{l}_{12}^\top \boldsymbol{s}_{3\parallel} = \boldsymbol{0} \vee  \boldsymbol{l}_5^\top \boldsymbol{s}_{1\parallel} = \boldsymbol{0}) $
\end{enumerate}

\subsubsection{The $5\$_0$ system limbs \texttt{Case 2}} 

In this scenario, where three revolute joints are parallel, the constraint equation is defined as follows:

\begin{equation}
	\begin{bmatrix}
		\boldsymbol{s}_{1\parallel} &  (\boldsymbol{s}_{1\parallel} \times \boldsymbol{l}_1)^\top \\ 
		\boldsymbol{s}_{1\parallel} &  (\boldsymbol{s}_{1\parallel} \times \boldsymbol{l}_2)^\top \\
		\boldsymbol{s}_{1\parallel} &  (\boldsymbol{s}_{1\parallel} \times \boldsymbol{l}_3)^\top \\
		\boldsymbol{s}_{4\parallel} &  (\boldsymbol{s}_{4\parallel} \times \boldsymbol{l}_4)^\top \\
		\boldsymbol{s}_{5\parallel} &  (\boldsymbol{s}_{5\parallel} \times \boldsymbol{l}_5)^\top \\
	\end{bmatrix}   \begin{bmatrix} \boldsymbol{s}_{r\perp} \\ \boldsymbol{s}_{r\parallel} \end{bmatrix} = \boldsymbol{0}  \label{eq:case I_con}
\end{equation}
where $\boldsymbol{s}_{5\parallel}$ must be orthogonal to $\boldsymbol{s}_{1\parallel}$ and $\boldsymbol{s}_{4\parallel}$ ensuring linear independence, hence $\boldsymbol{s}_{5\parallel} = \boldsymbol{s}_{1\parallel} \times \boldsymbol{s}_{4\parallel}$. Solving (\ref{eq:case I_con}), the limb restriction screw direction vector becomes 

I) $\boldsymbol{s}_{r\parallel} = \boldsymbol{s}_{1\parallel}$. To have a zero moment vector, $\begin{bmatrix} (\boldsymbol{s}_{1\parallel} \times \boldsymbol{l}_1)^\top \\ (\boldsymbol{s}_{1\parallel} \times \boldsymbol{l}_2)^\top \\ (\boldsymbol{s}_{1\parallel} \times \boldsymbol{l}_3)^\top \\ (\boldsymbol{s}_{4\parallel} \times \boldsymbol{l}_4)^\top \\ (\boldsymbol{s}_{5\parallel} \times \boldsymbol{l}_5)^\top \end{bmatrix} \boldsymbol{s}_{r\parallel} = \boldsymbol{C}_2 \boldsymbol{s}_{r\parallel} = \boldsymbol{0}$ is the condition.   

This condition can be satisfied if the followings holds true 

\begin{enumerate}
	\item  $\boldsymbol{s}_{4\parallel} \times \boldsymbol{l}_4 = 0 \vee \boldsymbol{s}_{1\parallel} \times \boldsymbol{l}_4 = 0 $, i.e., either $\boldsymbol{s}_{4\parallel} \parallel\boldsymbol{l}_4 $ or $\boldsymbol{l}_4  = 0 $, meaning the axis is coincident with the reference point.  
	\item  $ \boldsymbol{s}_{4\parallel}^\top\boldsymbol{l}_{5} = \boldsymbol{0} \wedge  ( \boldsymbol{s}_{4\parallel}^\top \boldsymbol{l}_{5} = \boldsymbol{0} \vee \boldsymbol{s}_{4\parallel}^\top \boldsymbol{s}_{1\parallel} = \boldsymbol{0})$  
\end{enumerate}

If these conditions are simultaneously satisfied, we get a restriction screw of non-zero direction and zero moment vectors for the $5\$_0$ system limb which in turn lead to a spherical parallel robot.

\subsection{The $4\$_0 - 1\$_\infty$ system limbs } 
In this specific limb configuration, the fourth revolute joint is orthogonal to one of the parallel joints and the third joint is positioned to ensure linear independence. Consequently, the constraint equation is formulated as follows

\begin{equation}
	\begin{bmatrix}
		\boldsymbol{s}_{1\parallel} ^\top &  (\boldsymbol{s}_{1\parallel} \times \boldsymbol{l}_1)^\top \\ 
		\boldsymbol{s}_{1\parallel} ^\top &  (\boldsymbol{s}_{1\parallel} \times \boldsymbol{l}_2)^\top \\ 
		\boldsymbol{s}_{3\parallel} ^\top &  (\boldsymbol{s}_{3\parallel} \times \boldsymbol{l}_3)^\top \\ 
		(\boldsymbol{s}_{1\parallel} \times \boldsymbol{s}_{3\parallel} ) ^\top &  \big((\boldsymbol{s}_{1\parallel} \times \boldsymbol{s}_{3\parallel}) \times \boldsymbol{l}_4\big)^\top \\ 
		\boldsymbol{0}^\top & \boldsymbol{s}_{5\parallel}^\top
	\end{bmatrix} 	\begin{bmatrix} \boldsymbol{s}_{r\perp}	\\ \boldsymbol{s}_{r\parallel}	 \end{bmatrix}  = \boldsymbol{0} 
\end{equation}

In this scenario, there are two sub-cases for the direction vector.

I) The direction vector $\boldsymbol{s}_{r\parallel} = \boldsymbol{s}_{1\parallel}$ with the condition $\boldsymbol{s}_{1\parallel}^\top \boldsymbol{s}_{5\parallel} = \boldsymbol{0} $ while the rank condition to have $\boldsymbol{s}_{r\perp} = \boldsymbol{0}$ is 

\begin{enumerate}
	\item  $ \boldsymbol{s}_{3\parallel} \times \boldsymbol{l}_3 = \boldsymbol{0} \vee \boldsymbol{s}_{1\parallel} \times \boldsymbol{l}_3 = \boldsymbol{0}$
	\item  $ \boldsymbol{l}_4^T\boldsymbol{s}_{3\parallel} = 0 \wedge (\boldsymbol{s}_{1\parallel}^\top \boldsymbol{l}_4 = 0 \vee \boldsymbol{s}_{3\parallel}^\top \boldsymbol{s}_{1\parallel}=0)$
\end{enumerate}

II) $\boldsymbol{s}_{1\parallel} = \boldsymbol{l}_{12} = \boldsymbol{0}$ with the condition $\boldsymbol{l}_{12}^\top \boldsymbol{s}_{5\parallel} = 0$. In this case, the moment vector $\boldsymbol{s}_{r\perp} = \boldsymbol{0} $ when the following conditions are simultaneously satisfied. 

\begin{enumerate}
	\item $\boldsymbol{l}_{1} \times \boldsymbol{s}_{1\parallel} = \boldsymbol{0} \vee \boldsymbol{s}_{1\parallel} \times \boldsymbol{l}_2 \vee \boldsymbol{l}_1 \times \boldsymbol{l}_2 = \boldsymbol{0}    $
	\item $ \boldsymbol{s}_{3\parallel}^\top \boldsymbol{l}_4 = 0 \wedge (\boldsymbol{s}_{1\parallel}^\top \boldsymbol{l}_4 = 0 \vee \boldsymbol{l}_3^\top \boldsymbol{l}_{12} = 0) $
	\item $ \boldsymbol{l}_{3} \times \boldsymbol{s}_{3\parallel} = \boldsymbol{0} \vee \boldsymbol{l}_{12} \times \boldsymbol{s}_{3\parallel} =  \boldsymbol{0} \vee \boldsymbol{l}_{12} \times \boldsymbol{l}_3 =\boldsymbol{0} $
\end{enumerate}

\subsection{$3\$_0 - 2\$_\infty$  system limbs}

\begin{equation}
	\begin{bmatrix}
		\boldsymbol{s}_{1\parallel} ^\top &  (\boldsymbol{s}_{1\parallel} \times \boldsymbol{l}_1)^\top \\ 
		\boldsymbol{s}_{2\parallel} ^\top &  (\boldsymbol{s}_{1\parallel} \times \boldsymbol{l}_2)^\top \\ 
		\boldsymbol{s}_{3\parallel} ^\top &  (\boldsymbol{s}_{3\parallel} \times \boldsymbol{l}_3)^\top \\ 
		\boldsymbol{0}^\top &  \boldsymbol{s}_{4\parallel}^\top \\ 
		\boldsymbol{0}^\top & \boldsymbol{s}_{5\parallel}^\top
	\end{bmatrix} 	\begin{bmatrix} \boldsymbol{s}_{r\perp}	\\ \boldsymbol{s}_{r\parallel}	 \end{bmatrix}  = \boldsymbol{0} 
\end{equation}

In any case, the direction vector is given by \(\boldsymbol{s}_{r\parallel} = \boldsymbol{s}_{4\parallel} \times \boldsymbol{s}_{5\parallel}\). Hence, for \(\boldsymbol{s}_{r\parallel} = \boldsymbol{0}\), it must be that \(\boldsymbol{s}_{4\parallel} \parallel \boldsymbol{s}_{5\parallel}\), implying the two prismatic joints should remain parallel. The moment vector is then derived from \(\boldsymbol{C}_1\boldsymbol{s}_{r\perp} = \boldsymbol{0}\). To ensure \(\boldsymbol{s}_{r\perp} \neq \boldsymbol{0}\), \(\boldsymbol{C}_1\) must be rank deficient, subject to the condition \(\boldsymbol{l}_1 = \boldsymbol{l}_2 = \boldsymbol{l}_3 = \boldsymbol{0}\). With this condition and through the duality relation, the zero moment vector is straightforwardly obtained, ensuring the necessary condition for \(SO(3)\) motion is satisfied.

\subsection{The $4\$_0$ system limbs}

\begin{equation}
	\begin{bmatrix}
		\boldsymbol{s}_{1\parallel}^\top &  (\boldsymbol{s}_{1\parallel} \times \boldsymbol{l}_1)^\top \\ 
		\boldsymbol{s}_{2\parallel}^\top &  (\boldsymbol{s}_{2\parallel} \times \boldsymbol{l}_2)^\top \\
		\boldsymbol{s}_{3\parallel}^\top &  (\boldsymbol{s}_{3\parallel} \times \boldsymbol{l}_3)^\top \\
		\boldsymbol{s}_{4\parallel}^\top &  (\boldsymbol{s}_{4\parallel} \times \boldsymbol{l}_4)^\top \\
	\end{bmatrix}   \begin{bmatrix} \boldsymbol{s}_{r\perp} \\ \boldsymbol{s}_{r\parallel} \end{bmatrix} = \boldsymbol{0}  \label{eq:four_sys_Case_1}
\end{equation}
This is the less complicated case and the condition for $\boldsymbol{s}_{r\perp}  =\boldsymbol{0}$ is $\boldsymbol{l}_1 = \boldsymbol{l}_2 = \boldsymbol{l}_3 = \boldsymbol{l}_4 = \boldsymbol{0}$ which also implies that all the direction vectors passes through the reference point (center of rotation). Hence, this is a redundant case. In Table \ref{tab:limbs_spm}, $\boldsymbol{s}_{\parallel}$ represents the direction of the joints, superscript $i$ indicates that these joints intersect at a common point, and $p$ signifies that the joints are parallel.      

\begin{table}[htb!]
	\caption{Limb variants for $SO(3)$ robots}
	\label{tab:limbs_spm}
	\setlength\tabcolsep{2pt}        % Adjust the column spacing as needed
	\renewcommand{\arraystretch}{1.5} % Adjust the row spacing as needed
	\begin{tabular}{|c|cc|}
		\hline
		Category & \multicolumn{2}{c|}{Type} \\ \hline
		\multirow{20}{*}{$5\$_0$} & \multicolumn{1}{c|}{\multirow{12}{*}{I}} & $  \boldsymbol{s}^i_{1\parallel}\boldsymbol{s}^i_{2\parallel}\boldsymbol{s}^i_{3\parallel}\boldsymbol{s}_{4\parallel}\boldsymbol{s}_{5\parallel}$ \\ \cline{3-3} 
		& \multicolumn{1}{c|}{} & $ \boldsymbol{s}^i_{1\parallel}\boldsymbol{s}^i_{2\parallel}\boldsymbol{s}_{3\parallel}\boldsymbol{s}_{4\parallel} \boldsymbol{s}_{5\parallel} $ \\ \cline{3-3} 
		& \multicolumn{1}{c|}{} & $ \boldsymbol{s}^i_{1\parallel}\boldsymbol{s}^i_{2\parallel}\boldsymbol{s}_{3\parallel} \boldsymbol{s}_{4\parallel} \boldsymbol{s}^i_{5\parallel} $ \\ \cline{3-3} 
		& \multicolumn{1}{c|}{} & $ \boldsymbol{s}^i_{1\parallel}\boldsymbol{s}^i_{2\parallel}\boldsymbol{s}_{3\parallel} \boldsymbol{s}^i_{4\parallel} \boldsymbol{s}_{5\parallel} $ \\ \cline{3-3} 
		& \multicolumn{1}{c|}{} & $ \boldsymbol{s}^i_{1\parallel}\boldsymbol{s}_{2\parallel}\boldsymbol{s}^i_{3\parallel}\boldsymbol{s}_{4\parallel} \boldsymbol{s}^i_{5\parallel} $ \\ \cline{3-3} 
		& \multicolumn{1}{c|}{} & $ \boldsymbol{s}^i_{1\parallel}\boldsymbol{s}_{2\parallel}\boldsymbol{s}_{3\parallel}\boldsymbol{s}^i_{4\parallel} \boldsymbol{s}^i_{5\parallel} $ \\ \cline{3-3} 
		& \multicolumn{1}{c|}{} & $ \boldsymbol{s}_{1\parallel}\boldsymbol{s}^i_{2\parallel}\boldsymbol{s}^i_{3\parallel} \boldsymbol{s}^i_{4\parallel} \boldsymbol{s}_{5\parallel} $ \\ \cline{3-3} 
		& \multicolumn{1}{c|}{} & $ \boldsymbol{s}_{1\parallel}\boldsymbol{s}^i_{2\parallel}\boldsymbol{s}^i_{3\parallel}\boldsymbol{s}_{4\parallel} \boldsymbol{s}^i_{5\parallel} $ \\ \cline{3-3} 
		& \multicolumn{1}{c|}{} & $ \boldsymbol{s}_{1\parallel}\boldsymbol{s}^i_{2\parallel}\boldsymbol{s}_{3\parallel}\boldsymbol{s}^i_{4\parallel} \boldsymbol{s}^i_{5\parallel} $ \\ \cline{3-3} 
		& \multicolumn{1}{c|}{} & $ \boldsymbol{s}_{1\parallel}\boldsymbol{s}_{2\parallel} \boldsymbol{s}^i_{3\parallel}\boldsymbol{s}^i_{4\parallel} \boldsymbol{s}^i_{5\parallel} $ \\ \cline{2-3} 
		%%%%%%%%%%%%%%%%%%%%%%%%%%%%%%%%%%%%%%%%%%%
		& \multicolumn{1}{c|}{\multirow{2}{*}{II}} & $\boldsymbol{s}^i_{1\parallel}\boldsymbol{s}^i_{2\parallel}\boldsymbol{s}^p_{3\parallel}  \boldsymbol{s}^p_{4\parallel}  \boldsymbol{s}^p_{5\parallel} $ \\ \cline{3-3} 
		& \multicolumn{1}{c|}{} &   $ \boldsymbol{s}^i_{1\parallel}\boldsymbol{s}^p_{2\parallel}\boldsymbol{s}^p_{3\parallel}\boldsymbol{s}^p_{4\parallel} \boldsymbol{s}^i_{5\parallel} $ \\ \cline{3-3} 
		& \multicolumn{1}{c|}{} &   $ \boldsymbol{s}^p_{1\parallel}\boldsymbol{s}^p_{2\parallel}\boldsymbol{s}^p_{3\parallel}\boldsymbol{s}^i_{4\parallel} \boldsymbol{s}^i_{5\parallel} $ \\ \cline{3-3}
		& \multicolumn{1}{c|}{} &   $ \boldsymbol{s}^i_{1\parallel}\boldsymbol{s}^i_{2\parallel}\boldsymbol{s}^p_{3\parallel}\boldsymbol{s}^p_{4\parallel} \boldsymbol{s}^p_{5\parallel} $ \\ \cline{3-3}
		& \multicolumn{1}{c|}{} &   $ \boldsymbol{s}^i_{1\parallel}\boldsymbol{s}^p_{2\parallel}\boldsymbol{s}^p_{3\parallel}\boldsymbol{s}^p_{4\parallel} \boldsymbol{s}^i_{5\parallel} $ \\ \cline{3-3}
		& \multicolumn{1}{c|}{} &   $ \boldsymbol{s}^p_{1\parallel}\boldsymbol{s}^p_{2\parallel}\boldsymbol{s}^p_{3\parallel}\boldsymbol{s}^i_{4\parallel} \boldsymbol{s}^i_{5\parallel} $ \\ \hline
		%%%%%%%%%%%%%%%%%%%%%%%%%%%%%%%%%%%%%%%%%%%
		\multirow{12}{*}{$4\$_0-1\$_\infty$} & \multicolumn{1}{c|}{I} & $ 20=\frac{5 !}{3!} ~\text{of} ~ \boldsymbol{s}^i_{1i\parallel}\boldsymbol{s}^i_{2i\parallel}\boldsymbol{s}^i_{3i\parallel}\boldsymbol{s}_{4i\parallel}\boldsymbol{s}_{5i\parallel} \mapsto \mathrm{RRRRP}$ \\ \cline{2-3}
		& \multicolumn{1}{c|}{\multirow{10}{*}{II}} &   $ \boldsymbol{s}^i_{1\parallel}\boldsymbol{s}^i_{2\parallel}\boldsymbol{s}^p_{3\parallel}\boldsymbol{s}^p_{4\parallel} \boldsymbol{s}^p_{5\parallel} \mapsto \mathrm{RRRRP}$ \\ \cline{3-3}
		& \multicolumn{1}{c|}{} &   $ \boldsymbol{s}^i_{1\parallel}\boldsymbol{s}^i_{2\parallel}\boldsymbol{s}^p_{3\parallel}\boldsymbol{s}^p_{4\parallel} \boldsymbol{s}^p_{5\parallel} \mapsto \mathrm{RRRPR} $   \\ \cline{3-3}
		& \multicolumn{1}{c|}{} &   $ \boldsymbol{s}^i_{1\parallel}\boldsymbol{s}^i_{2\parallel}\boldsymbol{s}^p_{3\parallel}\boldsymbol{s}^p_{4\parallel} \boldsymbol{s}^p_{5\parallel} \mapsto \mathrm{RRPRR} $   \\ \cline{3-3}
		& \multicolumn{1}{c|}{} &   $ \boldsymbol{s}^i_{1\parallel}\boldsymbol{s}^p_{2\parallel}\boldsymbol{s}^p_{3\parallel}\boldsymbol{s}^p_{4\parallel} \boldsymbol{s}^i_{5\parallel} \mapsto \mathrm{RRRPR} $   \\ \cline{3-3}
		& \multicolumn{1}{c|}{} &   $ \boldsymbol{s}^i_{1\parallel}\boldsymbol{s}^p_{2\parallel}\boldsymbol{s}^p_{3\parallel}\boldsymbol{s}^p_{4\parallel} \boldsymbol{s}^i_{5\parallel} \mapsto \mathrm{RRPRR} $   \\ \cline{3-3}
		& \multicolumn{1}{c|}{} &   $ \boldsymbol{s}^i_{1\parallel}\boldsymbol{s}^P_{2\parallel}\boldsymbol{s}^p_{3\parallel}\boldsymbol{s}^p_{4\parallel} \boldsymbol{s}^i_{5\parallel} \mapsto \mathrm{RPRRR} $   \\ \cline{3-3}
		& \multicolumn{1}{c|}{} &   $ \boldsymbol{s}^p_{1\parallel}\boldsymbol{s}^p_{2\parallel}\boldsymbol{s}^p_{3\parallel}\boldsymbol{s}^i_{4\parallel} \boldsymbol{s}^i_{5\parallel} \mapsto \mathrm{RRPRR} $   \\ \cline{3-3}
		& \multicolumn{1}{c|}{} &   $ \boldsymbol{s}^p_{1\parallel}\boldsymbol{s}^p_{2\parallel}\boldsymbol{s}^p_{3\parallel}\boldsymbol{s}^i_{4\parallel} \boldsymbol{s}^i_{5\parallel} \mapsto \mathrm{RPRRR} $   \\ \cline{3-3}
		& \multicolumn{1}{c|}{} &   $ \boldsymbol{s}^p_{1\parallel}\boldsymbol{s}^p_{2\parallel}\boldsymbol{s}^p_{3\parallel}\boldsymbol{s}^i_{4\parallel} \boldsymbol{s}^i_{5\parallel} \mapsto \mathrm{PRRRR} $   \\ \hline
		%%%%%%%%%%%%%%%%%%%%%%%%%%%%%%%%%%%%%%%%%%%
	\end{tabular}
\end{table}

\begin{table}[htb!]
%	\caption{Limb variants for $SO(3)$ robots}
	%\label{tab:limbs_spm}
	\setlength\tabcolsep{2pt}        % Adjust the column spacing as needed
	\renewcommand{\arraystretch}{1.5} % Adjust the row spacing as needed
	\begin{tabular}{|c|cc|}
		\hline
		Category & \multicolumn{2}{c|}{Type} \\ \hline		
		%%%%%%%%%%%%%%%%%%%%%%%%%%%%%%%%%%%%%%%%%%		
		\multirow{25}{*}{$3\$_0 -2\$_\infty$} & \multicolumn{1}{c|}{\multirow{13}{*}{I}} & $  \boldsymbol{s}^i_{1\parallel}\boldsymbol{s}^i_{2\parallel}\boldsymbol{s}^i_{3\parallel}\boldsymbol{s}_{4\parallel}\boldsymbol{s}_{5\parallel} \mapsto \mathrm{RRRPP} $ $\mapsto$ non $(\cdot)^i=$ category B\\ \cline{3-3}
		& \multicolumn{1}{c|}{} &   $ \boldsymbol{s}^i_{1\parallel}\boldsymbol{s}^i_{2\parallel}\boldsymbol{s}_{3\parallel}\boldsymbol{s}^i_{4\parallel} \boldsymbol{s}_{5\parallel}  \mapsto \mathrm{RRRPR}$ $\mapsto$ non $(\cdot)^i=$ category B\\ \cline{3-3} 
		& \multicolumn{1}{c|}{} &   $ \boldsymbol{s}^i_{1\parallel}\boldsymbol{s}^i_{2\parallel}\boldsymbol{s}_{3\parallel}\boldsymbol{s}_{4\parallel} \boldsymbol{s}^i_{5\parallel}  \mapsto \mathrm{RRPPR}$ $\mapsto$ non $(\cdot)^i=$ category B\\ \cline{3-3}
		& \multicolumn{1}{c|}{} &   $ \boldsymbol{s}^i_{1\parallel}\boldsymbol{s}_{2\parallel}\boldsymbol{s}^i_{3\parallel}\boldsymbol{s}^i_{4\parallel} \boldsymbol{s}_{5\parallel}  \mapsto \mathrm{RRRRP}$ $\mapsto$ non $(\cdot)^i=$ category B\\ \cline{3-3}
		& \multicolumn{1}{c|}{} &   $ \boldsymbol{s}^i_{1\parallel}\boldsymbol{s}_{2\parallel}\boldsymbol{s}^i_{3\parallel}\boldsymbol{s}_{4\parallel} \boldsymbol{s}^i_{5\parallel}  \mapsto \mathrm{RPRRP}$ $\mapsto$ non $(\cdot)^i=$ category B\\ \cline{3-3}
		& \multicolumn{1}{c|}{} &   $ \boldsymbol{s}^i_{1\parallel}\boldsymbol{s}_{2\parallel}\boldsymbol{s}_{3\parallel}\boldsymbol{s}^i_{4\parallel} \boldsymbol{s}^i_{5\parallel}  \mapsto \mathrm{RPRPR}$ $\mapsto$ non $(\cdot)^i=$ category B\\ \cline{3-3}
		& \multicolumn{1}{c|}{} &   $ \boldsymbol{s}_{1\parallel}\boldsymbol{s}^i_{2\parallel}\boldsymbol{s}^i_{3\parallel}\boldsymbol{s}^i_{4\parallel} \boldsymbol{s}_{5\parallel}  \mapsto \mathrm{PRRRP}$ $\mapsto$ non $(\cdot)^i=$ category B\\ \cline{3-3}
		& \multicolumn{1}{c|}{} &   $ \boldsymbol{s}_{1\parallel}\boldsymbol{s}^i_{2\parallel}\boldsymbol{s}^i_{3\parallel}\boldsymbol{s}_{4\parallel} \boldsymbol{s}^i_{5\parallel}  \mapsto \mathrm{PRRPR}$ $\mapsto$ non $(\cdot)^i=$ category B\\ \cline{3-3}
		& \multicolumn{1}{c|}{} &   $ \boldsymbol{s}_{1\parallel}\boldsymbol{s}^i_{2\parallel}\boldsymbol{s}_{3\parallel}\boldsymbol{s}^i_{4\parallel} \boldsymbol{s}^i_{5\parallel}  \mapsto \mathrm{PRPRR}$ $\mapsto$ non $(\cdot)^i=$ category B\\ \cline{3-3}
		& \multicolumn{1}{c|}{} &   $ \boldsymbol{s}_{1\parallel}\boldsymbol{s}_{2\parallel}\boldsymbol{s}^i_{3\parallel}\boldsymbol{s}^i_{4\parallel} \boldsymbol{s}^i_{5\parallel}  \mapsto \mathrm{PPRRR}$ $\mapsto$ non $(\cdot)^i=$ category B\\ \cline{2-3}
		%%%%%
		& \multicolumn{1}{c|}{\multirow{12}{*}{II}} &   $ \boldsymbol{s}^i_{1\parallel}\boldsymbol{s}^i_{2\parallel}\boldsymbol{s}^p_{3\parallel}\boldsymbol{s}^p_{4\parallel} \boldsymbol{s}^p_{5\parallel} \mapsto \mathrm{RRRPP}  $  $\mapsto$ non $(\cdot)^i=$ category B \\ \cline{3-3}
		& \multicolumn{1}{c|}{} &   $ \boldsymbol{s}^i_{1\parallel}\boldsymbol{s}^i_{2\parallel}\boldsymbol{s}^p_{3\parallel}\boldsymbol{s}^p_{4\parallel} \boldsymbol{s}^p_{5\parallel} \mapsto \mathrm{RRPRP} $ $\mapsto$ non $(\cdot)^i=$ category B \\ \cline{3-3} 
		& \multicolumn{1}{c|}{} &   $ \boldsymbol{s}^i_{1\parallel}\boldsymbol{s}^i_{2\parallel}\boldsymbol{s}^p_{3\parallel}\boldsymbol{s}^p_{4\parallel} \boldsymbol{s}^p_{5\parallel} \mapsto \mathrm{RRPPR} $ $\mapsto$ non $(\cdot)^i=$ category B\\ \cline{3-3}
		& \multicolumn{1}{c|}{} &   $ \boldsymbol{s}^i_{1\parallel}\boldsymbol{s}^p_{2\parallel}\boldsymbol{s}^p_{3\parallel}\boldsymbol{s}^p_{4\parallel} \boldsymbol{s}^i_{5\parallel} \mapsto \mathrm{RRPPR} $ $\mapsto$ non $(\cdot)^i=$ category B \\ \cline{3-3}
		& \multicolumn{1}{c|}{} &   $ \boldsymbol{s}^i_{1\parallel}\boldsymbol{s}^p_{2\parallel}\boldsymbol{s}^p_{3\parallel}\boldsymbol{s}^p_{4\parallel} \boldsymbol{s}^i_{5\parallel} \mapsto \mathrm{RPRPR} $ $\mapsto$ non $(\cdot)^i=$ category B \\ \cline{3-3}
		& \multicolumn{1}{c|}{} &   $ \boldsymbol{s}^i_{1\parallel}\boldsymbol{s}^p_{2\parallel}\boldsymbol{s}^p_{3\parallel}\boldsymbol{s}^p_{4\parallel} \boldsymbol{s}^i_{5\parallel} \mapsto \mathrm{RPPRR} $ $\mapsto$ non $(\cdot)^i=$ category B \\ \cline{3-3}
		& \multicolumn{1}{c|}{} &   $ \boldsymbol{s}^P_{1\parallel}\boldsymbol{s}^p_{2\parallel}\boldsymbol{s}^p_{3\parallel}\boldsymbol{s}^i_{4\parallel} \boldsymbol{s}^i_{5\parallel} \mapsto \mathrm{RPPRR} $ $\mapsto$ non $(\cdot)^i=$ category B \\ \cline{3-3}
		& \multicolumn{1}{c|}{} &   $ \boldsymbol{s}^P_{1\parallel}\boldsymbol{s}^p_{2\parallel}\boldsymbol{s}^p_{3\parallel}\boldsymbol{s}^i_{4\parallel} \boldsymbol{s}^i_{5\parallel} \mapsto \mathrm{PRPRR} $ $\mapsto$ non $(\cdot)^i=$ category B \\ \cline{3-3}
		& \multicolumn{1}{c|}{} &   $ \boldsymbol{s}^P_{1\parallel}\boldsymbol{s}^p_{2\parallel}\boldsymbol{s}^p_{3\parallel}\boldsymbol{s}^i_{4\parallel} \boldsymbol{s}^i_{5\parallel} \mapsto \mathrm{PPRRR} $ $\mapsto$ non $(\cdot)^i=$ category B\\ \hline
		%%%%%%%%%%%%%%%%%%%%%%%%%%%%%%%%%%%
	\end{tabular}
\end{table}

\begin{table}[htb!]
%	\caption{ Limb variants for $SO(3)$ robots}
	%\label{tab:limbs_spma}
	\setlength\tabcolsep{2pt}        % Adjust the column spacing as needed
	\renewcommand{\arraystretch}{1.5} % Adjust the row spacing as needed
	\begin{tabular}{|c|cc|}
		\hline
		Category & \multicolumn{2}{c|}{Type} \\ \hline		
		\multirow{4}{*}{$4\$_0 -0\$_\infty$} & \multicolumn{1}{c|}{\multirow{4}{*}{I}} &   $ \boldsymbol{s}^i_{1\parallel}\boldsymbol{s}^i_{2\parallel}\boldsymbol{s}^i_{3\parallel}\boldsymbol{s}_{4\parallel}   \mapsto \mathrm{RRRR}  $  $\mapsto$ redundant case \\ \cline{3-3} 
		& \multicolumn{1}{c|}{\multirow{4}{*}{}} &   $ \boldsymbol{s}^i_{1\parallel}\boldsymbol{s}^i_{2\parallel}\boldsymbol{s}_{3\parallel}\boldsymbol{s}^i_{4\parallel}    \mapsto \mathrm{RRRR}  $  $\mapsto$ redundant case \\ \cline{3-3} 
		& \multicolumn{1}{c|}{\multirow{4}{*}{}} &   $ \boldsymbol{s}^i_{1\parallel}\boldsymbol{s}_{2\parallel}\boldsymbol{s}^i_{3\parallel}\boldsymbol{s}^i_{4\parallel}  \mapsto \mathrm{RRRR}  $  $\mapsto$ redundant case \\ \cline{3-3} 
		& \multicolumn{1}{c|}{\multirow{4}{*}{}} &   $ \boldsymbol{s}_{1\parallel}\boldsymbol{s}^i_{2\parallel}\boldsymbol{s}^i_{3\parallel}\boldsymbol{s}^i_{4\parallel}  \mapsto \mathrm{RRRR}  $  $\mapsto$ redundant case \\ \hline
		
		\multirow{4}{*}{$3\$_0 -1\$_\infty$} & \multicolumn{1}{c|}{\multirow{4}{*}{II}} &   $ \boldsymbol{s}^i_{1\parallel}\boldsymbol{s}^i_{2\parallel}\boldsymbol{s}^i_{3\parallel}\boldsymbol{s}_{4\parallel}  \mapsto \mathrm{RRRP}  $   $\mapsto$ $\$_\infty$ no effect\\ \cline{3-3} 
		& \multicolumn{1}{c|}{\multirow{8}{*}{}} &   $ \boldsymbol{s}^i_{1\parallel}\boldsymbol{s}^i_{2\parallel}\boldsymbol{s}_{3\parallel}\boldsymbol{s}^i_{4\parallel}   \mapsto \mathrm{RRRP}  $  $\mapsto$ $\$_\infty$ no effect \\ \cline{3-3} 
		& \multicolumn{1}{c|}{\multirow{8}{*}{}} &   $ \boldsymbol{s}^i_{1\parallel}\boldsymbol{s}_{2\parallel}\boldsymbol{s}^i_{3\parallel}\boldsymbol{s}^i_{4\parallel}   \mapsto \mathrm{RRRP}  $  $\mapsto$ $\$_\infty$ no effect \\ \cline{3-3} 
		& \multicolumn{1}{c|}{\multirow{8}{*}{}} &   $ \boldsymbol{s}_{1\parallel}\boldsymbol{s}^i_{2\parallel}\boldsymbol{s}^i_{3\parallel}\boldsymbol{s}^i_{4\parallel}   \mapsto \mathrm{RRRP}  $  $\mapsto$ $\$_\infty$ no effect \\ \hline 
		\multirow{1}{*}{$3\$_0 -0\$_\infty$} & \multicolumn{1}{c|}{\multirow{1}{*}{I}} &   $ \boldsymbol{s}^i_{1\parallel}\boldsymbol{s}^i_{2\parallel}\boldsymbol{s}^i_{3\parallel}  \mapsto \mathrm{RRR}  $ $\mapsto$ typical case   \\ \hline
	\end{tabular}
\end{table}

\subsection{The $3\$_0 - 1\$_\infty$ system limbs}

\begin{equation}
	\begin{bmatrix}
		\boldsymbol{s}_{1\parallel}^\top &  (\boldsymbol{s}_{1\parallel} \times \boldsymbol{l}_1)^\top \\ 
		\boldsymbol{s}_{2\parallel}^\top &  (\boldsymbol{s}_{2\parallel} \times \boldsymbol{l}_2)^\top \\
		\boldsymbol{s}_{3\parallel}^\top &  (\boldsymbol{s}_{3\parallel} \times \boldsymbol{l}_3)^\top \\
		\boldsymbol{0}^\top &  \boldsymbol{s}_{4\parallel}^\top \\
	\end{bmatrix}   \begin{bmatrix} \boldsymbol{s}_{r\perp} \\ \boldsymbol{s}_{r\parallel} \end{bmatrix} = \boldsymbol{0}  \label{eq:four_sys_Case_2}
\end{equation}

The only condition to ensure both direction vectors in \(\boldsymbol{s}_{r\parallel} = \boldsymbol{0}\) is \(\boldsymbol{l}_1 = \boldsymbol{l}_2 = \boldsymbol{l}_3 = \boldsymbol{0}\). This occurs when all the revolute joint axes intersect at a single point, fulfilling the necessary condition for \(SO(3)\) motion. Furthermore, as discussed in Section \ref{sec:generic conditions}, the presence of a \(\$_\infty\) joint does not influence the motion generated by \(\$_0\) joints.

\subsection{The $3\$_0$ system limbs}

For this limb types, the general constraint equation becomes 

\begin{equation}
	\begin{bmatrix}
		\boldsymbol{s}_{1\parallel}^\top & (\boldsymbol{s}_{1\parallel} \times \boldsymbol{l}_1)^\top  \\ 
		\boldsymbol{s}_{2\parallel}^\top & (\boldsymbol{s}_{2\parallel} \times \boldsymbol{l}_2)^\top  \\ 
		\boldsymbol{s}_{3\parallel}^\top & (\boldsymbol{s}_{3\parallel} \times \boldsymbol{l}_3)^\top  \\ 
	\end{bmatrix} \begin{bmatrix} \boldsymbol{s}_{r\perp} \\ \boldsymbol{s}_{r\parallel} \end{bmatrix} = \boldsymbol{0} \label{eq:3syste}
\end{equation}
The primary condition for $\boldsymbol{s}_{r\perp}$ is the independence of $\boldsymbol{s}_{1\parallel}, \boldsymbol{s}_{2\parallel}$, and $\boldsymbol{s}_{3\parallel}$, represented by $\begin{bmatrix} (\boldsymbol{s}_{1\parallel} \times \boldsymbol{l}_1)^\top \ (\boldsymbol{s}_{2\parallel} \times \boldsymbol{l}_2)^\top \ (\boldsymbol{s}_{3\parallel} \times \boldsymbol{l}_3)^\top \end{bmatrix} \boldsymbol{s}_{r\parallel} = \boldsymbol{C}_2^\top \boldsymbol{s}{r\parallel} = \boldsymbol{0}$. To achieve a non-trivial direction vector, $\boldsymbol{s}_{r\parallel} \neq \boldsymbol{0}$, as specified in (\ref{eq:3syste}), $\boldsymbol{C}_2$ must be rank deficient. This condition ultimately requires that all the joints intersect at a common point, as categorized in category A. 

The sufficient conditions for all cases, as described in (\ref{eq:rik_gen_con}), (\ref{eq:frk_gen_con}), (\ref{eq:rik_gen_con_final}), and (\ref{eq:frk_gen_con_final}), involve setting the matrix $\boldsymbol{G}_{av} = \boldsymbol{0}$, which is spanned by the direction vector of the actuation wrench. In this context, selecting the appropriate actuator is crucial, as different selections yield different results. However, this step becomes straightforward once the direction and moment vectors of the limb restriction screw are determined. According to the conditions outlined so far, all limb variants are enumerated in Table {\ref{tab:limbs_spm}}. As illustrated in the table, we have identified 73 distinct limb variants. This identification enables the construction of 73 symmetric spherical parallel robots, while the remaining 5256 variants can be achieved through various combinations of these limbs.

\section{Conclusions} \label{sec:conclusions}
In this work, addressing the limitations inherent in the case-by-case analysis of $SO(3)$ type parallel robots, also known as spherical robots, we have introduced a comprehensive array of limbs potentially suitable for this category of parallel robots, thereby unveiling all variants. Our approach utilized a purely analytical and algebraic procedure to derive generic and intuitive conditions for both the limbs and the entire manipulators. Initially, we analyzed the inverse Jacobian by partitioning it into block matrices to understand the most generic conditions for $SO(3)$ motion, guiding us toward detailed condition derivation. Subsequently, leveraging geometric characteristics of $SO(3)$ motion, we established specific criteria. We then thoroughly examined the relationships between direction and moment vectors, as well as position vectors, to uncover all conditions enabling $SO(3)$ motion generation. This exploration led to the identification of a complete array of limbs. Our findings reveal that there are 84 distinct variants of limbs capable of generating $SO(3)$ motion when assembled. In total, this suggests the possibility of constructing 84 symmetric and 6,972 asymmetric spherical robots, culminating in 7,056 $SO(3)$ spherical robots. This significantly broadens our options for selecting the most suitable and optimal robot for specific applications.

%\begin{table}[htbp]
%\caption{Table Type Styles}
%\begin{center}
%\begin{tabular}{|c|c|c|c|}
%\hline
%\textbf{Table}&\multicolumn{3}{|c|}{\textbf{Table Column Head}} \\
%\cline{2-4} 
%\textbf{Head} & \textbf{\textit{Table column subhead}}& \textbf{\textit{Subhead}}& \textbf{\textit{Subhead}} \\
%\hline
%copy& More table copy$^{\mathrm{a}}$& &  \\
%\hline
%\multicolumn{4}{l}{$^{\mathrm{a}}$Sample of a Table footnote.}
%\end{tabular}
%\label{tab1}
%\end{center}
%\end{table}

%\begin{figure}[htb!]
%	\centering
%	\includegraphics[width=\columnwidth]{spherical.pdf} % Adjusts the image to span the column width
%	\caption{Typical examples of spherical parallel manipulator obtained from Table \ref{tab2}.}
%	\label{fig:spherical}
%\end{figure}

\section*{Acknowledgment}

This research was supported by the Robotics Research Center of Yuyao (Grant No. KZ22308), National Natural Science Foundation of China under the Youth Program (Grant No. 509109-N72401) and the 2023 National High-level Talent Project, also within the Youth Program (Grant No. 588020-X42306/008).

\nocite{*}
\bibliographystyle{IEEEtran}
\bibliography{IEEEabrv,iros}

\end{document}